\documentclass[review]{elsarticle}

\usepackage{lineno,hyperref}
\usepackage{amsmath}
\modulolinenumbers[5]
\usepackage{color}
\usepackage{times}
\usepackage[T1]{fontenc}
\usepackage{subfig}
\usepackage{graphicx}
\usepackage{amsfonts}
\usepackage{amsmath}
\usepackage{color}
\usepackage{float}
\usepackage[usenames,dvipsnames,svgnames,table]{xcolor}
\usepackage{graphics}
\usepackage{epstopdf}
\DeclareMathOperator*{\argmax}{arg\,max}

\journal{Extended Bag-of-Tracklets for Multi-Face Tracking in Egocentric Photo-streams}

\begin{document}

\begin{frontmatter}

\title{Multi-Face Tracking by Extended Bag-of-Tracklets in Egocentric Videos}


\author[A]{Maedeh Aghaei \corref{cor1}}
\author[A,B]{Mariella Dimiccoli}
\author[A,B]{Petia Radeva}

\address[A]{University of Barcelona, MAIA Department, Barcelona, Spain}
\address[B]{Computer Vision Center, Cerdanyola del Vall{\'e}s (Barcelona), Spain}


\begin{abstract}
Wearable cameras offer a hands-free way to record egocentric images of daily experiences, where social events are of special interest. The first step towards detection  of social events is to track the appearance of multiple persons involved in it.
In this paper, we propose a novel method to find correspondences of multiple faces in low temporal resolution egocentric videos acquired through a wearable camera. This kind of photo-stream imposes additional challenges to the multi-tracking problem with respect to conventional videos.
Due to the free motion of the camera and to its low temporal resolution, abrupt changes in the field of view, in illumination condition and in  the target location are highly frequent. To overcome such difficulties, we propose a  multi-face tracking method that  generates a set of \textit{tracklets} through finding correspondences along the whole sequence for each detected face and takes advantage of the tracklets redundancy to deal with unreliable ones. Similar tracklets are grouped into the so called \textit{extended bag-of-tracklets} (eBoT), which is aimed to correspond to a specific person. Finally, a prototype tracklet is extracted for each eBoT, where the occurred occlusions are estimated by relying on a new measure of confidence. We validated our approach over an extensive dataset of egocentric photo-streams and compared it to state of the art methods, demonstrating its effectiveness and robustness.
\end{abstract}


\end{frontmatter}

\section{Introduction}
 
Wearable cameras and egocentric vision are very recent trends from barely the last ten years that paved the road for very challenging applications ranging from healthcare to sport, security, tourism, and leisure. Wearable cameras from the first person point of view record where a person is, what a person does and whom he/she interacts with. Thus, egocentric images are potentially useful for understanding the lifestyle of a person or quantified self. Egocentric images may also serve as digital memories, being particularly suited to boost the memory capabilities of people with memory impairment \cite{sensecam, sensecam1}. For this particular application, low temporal resolution wearable cameras, such as the Vicon Revue (3fpm) and the Narrative Clip (2fpm), are especially suited, since they allow the recording of one's life over a long period of time. However, extracting relevant information from egocentric videos with low temporal resolution, hereafter called \textit{egocentric photo-streams}, is not a trivial task. Indeed, a massive number of unconstrained collected images can be gathered even over relatively limited period of time (up to 3000 images per day using the Narrative Clip). Moreover, given the unpredictability of the camera motion and the low temporal resolution of the camera, abrupt changes of scene occur very often in the images.

\begin{figure}
\begin{center}
\includegraphics[width=12.5cm, height=1.5cm]{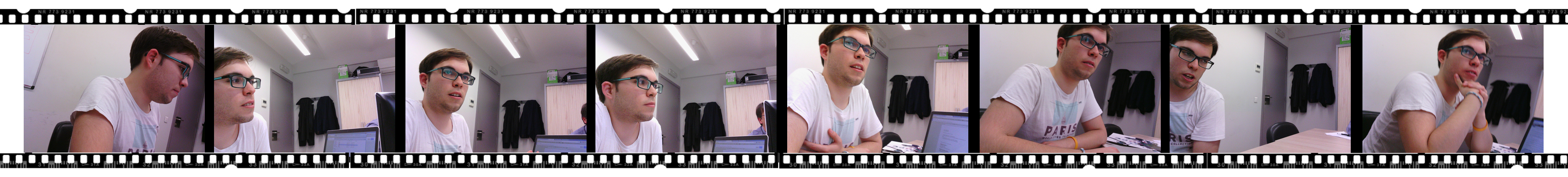}
\caption{A sequence of images acquired by Narrative clip wearable camera. Free motion of the camera and the abrupt variation in appearance of target due to low temporal resolution of the sensor can be appreciated.}
\label{fig1}
\end{center}
\end{figure}

During the last few years, several problems related to the analysis and organization of egocentric videos have been addressed, from temporal segmentation \cite{tempSeg, Tavalera2015} and summarization \cite{storyDriven, attenBased, egoActivity} to event and action (self action and social interaction) recognition \cite{action1, dailyLiving,socialInteraction,action2}. However, despite the importance of \emph{tracking} in the analysis of social interaction, this problem received less attention in egocentric vision than the same problem in conventional videos that has been an active research area for a long time \cite{survey}. Tracking in egocentric videos and in the special case of them, egocentric photo-streams, is a different problem from the tracking in conventional videos in several aspects. Conventional tracking facilitates itself with the assumption of temporal coherence, while temporal coherence does not hold for egocentric photo-streams. Moreover, in egocentric photo-streams, the appearance of the target as well as its position may change drastically from frame to frame. In addition, due to changes in the camera field of view caused by body movement of the camera wearer, background modeling becomes a more challenging issue (see Fig. \ref{fig1}).

When reviewing the state of the art trackers, two main categories of conventional trackers can be found: offline and online trackers. The former category of trackers assumes that object detection in all frames has already been performed and trajectory construction is achieved by linking different detections and tracks in offline mode \cite{kshortest, merged, GMCP}. This property of offline trackers allows for global optimization of the path and thus, makes them potentially suitable for dealing with photo-streams.
Berclaz et al. \cite{kshortest} reformulate the linking step between detections and trajectories as a constrained flow optimization approach, which results in a convex problem that can be solved using the k-shortest paths algorithm. In order to overcome the noisy probabilities of candidates that may be produced by the object detector, the authors arranged a set of assumptions including the limited motion of the target.
Zamir et al. \cite{GMCP} solve the data association problem for one object at a time, while implicitly incorporating the rest of the objects using global association by employing \textit{Generalized Minimum Clique Graphs} (GMCP). GMCP incorporates both motion and appearance model over the whole temporal span for optimization.
In the development of aforementioned trackers, the authors assume a rather fixed or predictable position for targets in the adjacent frames of the video. Although this assumption is generally applicable in conventional videos, it does not hold in egocentric photo-streams.

In comparison with offline trackers, the target position is provided for  online trackers in the initial frame and the tracker needs to establish the state of the target in the following frames of the video.
Among state of the art online trackers, those that are relatively tolerant to occlusion and drastic appearance changes, are more suitable for egocentric photo-streams \cite{CT, SPT, LOT, L1O}.
Kalal et al. present a \textit{Tracking, Learning, Detection} (TLD) framework \cite{TLD}, which works by training a discriminative classifier over labeled and unlabeled examples. This method performs well in handling short-term occlusions, but strongly relies on optical flow, which cannot be applied in low temporal resolution sequences.
\textit{Compressive Tracking} (CT) \cite{CT}, uses an appearance model based on features extracted in a compressed sensing domain. This method is relatively robust to changes in appearance and performs quite well in challenging datasets, outperforming TLD. However, CT is not robust to large displacements of the target, which are very frequent in egocentric sequences.
In \textit{Locally Orderless Tracking} (LOT) \cite{LOT}, target and candidates in the new frame are segmented first into superpixels and among the set of candidates, the one which has less distance to the target is selected as the target in the new frame. LOT tracker offers adaptation to object appearance variations by matching with flexible rigidity through measuring the distance between superpixels.
Similar to LOT, \textit{SuperPixel Tracker} (SPT) \cite{SPT} extracts superpixels of the target. SPT extracts the color histograms of the superpixels from the first 4 frames and based on these features, clusters superpixels by using mean-shift. A confidence value is assigned to each cluster, from which the superpixels confidence of all pixels of the cluster is derived. In the next frame, the candidate window with the highest confidence summed over all superpixels in the window is selected as the new target. Mei et al. presented L1O \cite{L1O} as a tracker which explicitly detects occlusions. In L1O, the candidate windows with a reconstruction error above a threshold are selected for L1-minimization.
When certain amount of the pixels of the candidate window are occluded, L1O detects an occlusion, which disables the model updating.

Conventional online trackers usually search for the target in the new frame, around its previous position in the current frame. These trackers are mostly dependent on the object appearance in the very first frames and generally require the feature patches in neighboring frames to be close to each other. However, under specific conditions of egocentric photo-streams, such presumptions will result in gradual departure of estimated target from the true target state, which eventually leads to tracking loss.

The work that seems the most similar to ours are the trackers in \textit{Low Frame Rate} (LFR) videos \cite{LFR1, NNLFR}. Li et al. present a temporal probabilistic combination of discriminative models of different learning and service period, known as their \textit{lifespan} \cite{LFR1}. Each model is learned from different ranges of samples, with different subsets of features, to achieve varying levels of discriminative power.
Different models are fused by a cascade particle filter, to achieve multiple stages of importance sampling. However, this work falls into pre-trained tracking class that  its performance also depends on the training data, an issue that we try to avoid, due to the peculiarity of our dataset that presents a relatively small number of images in each trackable segment.
A recent work about LFR tracking was presented by Zhou et al. \cite{NNLFR}. The authors proposed a \textit{Nearest Neighbor Field} (NNF) driven stochastic sampling framework for abrupt motion tracking. In this work, NNF provides candidate regions, where the target may exist. Smoothing Stochastic Approximate Monte Carlo (SSAMC) sampling scheme predicts the state of the target more effectively. Finally, the method refined the result with a sparse representation based template matching technique. 

Although the body of literature regarding tracking is huge, most existing approaches cannot be directly applied to egocentric photo-streams, either because of the unpredictability of motion or because of drastic appearance changes that characterize this data. Furthermore, most of the methods are not able to track multiple targets simultaneously or require the manual specification of the initial position of the target. To this end, we previously proposed the \textit{Bag-of-Tracklets} (BoT) \cite{BOT} for tracking in egocentric photo-streams acquired by Sensecam camera (3 fpm). The underlying key idea of our approach is that detection and tracking can be integrated to achieve strong discriminative power. This approach belongs to the offline class of trackers, that allows for general optimization of tracklets. Optimization consists of generating a tracklet for each detected target and categorizing similar tracklets into groups, that should correspond to different persons. This approach simply allows for rejection of unreliable bag-of-tracklets, and eventually extracts a single prototype for each reliable bag-of-tracklets. 

In this manuscript, we present an \textit{extended-Bag-of-Tracklets} (eBoT) approach by introducing several features that help in increasing BoT robustness even in photo-streams acquired by cameras with lower frame rates (2 fpm) and narrower fields of view:
\begin{itemize}

\item To manage the close appearance of people to the camera, eBoT reliably detects people characterizing them by their face instead of their body.

\item eBoT to handle target deformations and scale variations, employs a new approach for finding correspondences based on an average deep matching score. 

\item eBoT presents a more robust way to compute the prototype of the bag of tracklets.

\item eBoT is tolerant of face occlusions and is able to explicitly localize them.

\item eBoT introduces a confidence term to measure the reliability of the prototypes.

\item eBoT is compared to six models of the state of the art by using an enlarged set of metrics from the CLEAR MOT \cite{metrics} framework over an enlarged dataset.

\end{itemize}

The rest of the paper is organized as follows: in Sec. \ref{eBOT}, we define the Confidence-based eBoT for multi-face tracking, by performing seed and tracklet generation, grouping tracklets, prototypes extraction and occlusion treatment. In Sec. \ref{Results}, we introduce our experimental setup and discuss comparative results and finally, in Sec. \ref{sec:conclusions}, we end the paper by drawing conclusions and sketching future work.


\section{Confidence-based extended Bag-of-Tracklets for multi-face tracking}
\label{eBOT}
People during a full day may often engage in a social event. A social event typically happens in a specific environment with specific people. Thus, by wearing a wearable camera one captures those specific moments that are of interest for later retrieval. However, the first step towards social event retrieval from images is to find and track the appearance of people around the camera wearer. Precisely, people who get engaged in a social event with the camera wearer appear in reasonable number of consecutive frames while irrelevant people to the camera wearer only appear occasionally in the photos and normally do not stay in front of the camera wearer for a long time. Thus, by incorporating additional information about the tracked people, their involvement in the social interaction with the user can be proved \cite{aghaei2015towards}. Our approach to track multiple faces in egocentric photo-streams consists of four main steps: \emph{seed and tracklet generation}, \emph{grouping tracklets into Bag-of-tracklets}, \emph{prototypes extraction} and \emph{occlusion treatment}.


\subsection{Seed and Tracklet Generation}
Prior to any computation, the first step of the proposed method is to organize the long and unconstrained egocentric photo-streams into homogeneous temporal segments. To this end, we apply R-clustering \cite{Tavalera2015}, an unsupervised temporal segmentation method, specifically formulated for egocentric photo-streams. R-clustering consists in a Graph-Cut algorithm that finds a trade-off between the under-segmentation produced by a concept drift detector, and the over-segmentation resulting from agglomerative clustering. The clustering is performed over global features of images extracted through Convolutional Neural Networks to divide the photo-streams into  structured segments.

\begin{figure}[h]
\begin{center}
\includegraphics[width=11.8 cm, height=1.6 cm]{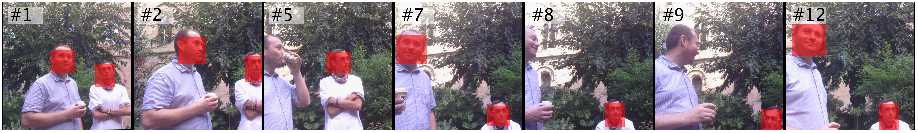}
\caption{Detected faces (seeds) are shown by red bounding boxes. An example of false negatives can be observed in frames 8 and 9. Only a sub-sample of the original sequence is shown.}
\label{seed}
\end{center}
\end{figure}

Among the set of created segments from the temporal segmentation step, those that contain trackable persons are of particular interest for our purpose. To determine if a segment contains trackable persons, we evaluate the  ratio between the number of frames with detected faces and the number of frames of the segment. If the ratio is higher than a predefined threshold (0.5 in this work), then the segment is considered as a segment containing trackable persons. As output of this phase, we collect a set of bounding boxes that surround the face of each person throughout the sequence, that we call \textit{seeds}. The generated seeds are shown by red bounding boxes in  Fig. \ref{seed}.

Due to the nature of our photos,  an \textit{in the wild} face detector \cite{faceDetection} that substantially outperforms state of the art face detectors \cite{viola}, is applied on each frame of the extracted segments to detect visible faces. The detector is based on mixture of trees with a shared pool of parts, where, every facial landmark is defined as a \textit{part} and a global mixtures is used to model topological changes due to the viewpoint. Different mixtures share part templates that allows modeling a large number of views with low complexity. Moreover, as shown by the authors, tree-structured models perform effectively at capturing global elastic deformation, while being easy to optimize using dynamic programming. Global mixtures can also be used to capture large deformation changes for a single viewpoint, such as changes in expression. Despite the relatively good performance of the detector, it sometimes produces some false positives or false negatives due to the blurring effect that happens frequently in egocentric photos.

Hereafter, we denote each segment containing trackable persons simply by \textit{sequence}. 
For each seed, we generate a set of correspondences to the seed along the sequence, called \textit{tracklet}, by propagating the seed in the sequence forward and backward using a similarity measure to be detailed below. As a result, a tracklet $T^i = \lbrace t_b^i,...,t^i_s,...,t_e^i\rbrace$ associated to the seed $i$ found at time $s$ begins in a time $b$,  where the backward tracking ends (first frame in the sequence), and ends at time $e$, where the forward tracking ends (last frame in the sequence). In the rest of the paper, we will keep the convention of using the variable $t$ to refer to the bounding box surrounding the faces, the upper-index to identify the tracklet, and the sub-index to identify the frame. Note that theoretically, the number of generated tracklets should be of order of the number of found seeds. For example, in the ideal case where face detector does not fail, two persons appearing in all the 100 frames of the sequence, would generate 200 tracklets, each one of length 100 frames.

To propagate a seed found in frame $s$, backward and forward, we look at every frame of the sequence to the region most similar to the seed. In order to deal with abrupt displacements of the target,  we generate the set of sample regions with a sliding window. However this approach generate a very high number of samples for each image, to reduce computational complexity, we reject all samples whose similarity to the seed in the HSV color space is lower than a pre-defined threshold. The size of the sliding window depends on the size of the seed that we are considering. However, since the face region in each frame can vary largely by its distance variation from the camera, we also consider as samples all previously detected seeds in that frame. 

The similarity between the seed and each sample in a frame of the sequence is measured by its average deep-matching score \cite{deepMatching}. The deep matching is conceived as a 2D-warping, that is able to deal with various kinds of object-induced or camera-induced image deformations, including scaling factors and rotations. Instead of using SIFT patches as descriptors, each SIFT patch is split into four so-called quadrants and, assuming independent motion (within some extent) of each of the four quadrants,  the similarity is computed to optimize the positions of the four quadrants of the target descriptor. 

\begin{figure}[h]
\begin{center}
\includegraphics[width=11.8cm, height=1.6cm]{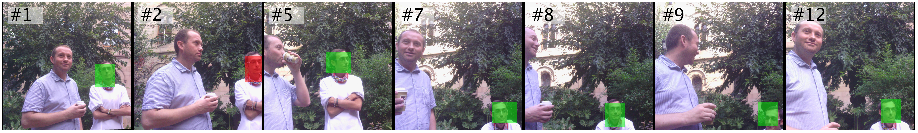}
\caption{An example of a tracklet generated based on deep matching. The red box corresponds to the seed that the tracklet is generated from it. The green box in each frame corresponds to the sample with the highest deep matching score to the seed.}
\label{tracklet}
\end{center}
\end{figure}

For simplicity, let us consider two sequences of R-dimensional descriptors in a 1D warping case:  the \textit{reference}, that corresponds to the seed,  say $P_s = {\left \{ p_{s,i} \right \}}_{i=0}^{R-1}$, and the \textit{target}, say $P_t= {\left \{ p_{t,i} \right \}}_{i=0}^{R-1}$, that corresponds to a sample in a frame. The optimal warping between them is defined by the function $w^*$ : $\left \{ 0,\ldots,R-1 \right \}\rightarrow \left \{ 0,\ldots,R-1 \right \}$ that maximizes the average value of similarities between their elements: 
\begin{equation}
\Lambda(w^*) = \max_{w\in W} S(w) = \max_{w\in W} M_i \lbrace sim (P_s(i),P_t(w(i))) \rbrace_{i = 0,\ldots,R-1}
\label{deepscore}
\end{equation}
where $w(i)$ returns the position of element $i$ in $P_t$, $M_i$ is the average value of the set of similarity values generate by varying $i$ and  $sim$ is the non-negative cosine similarity between pixel gradients. 
The deep matching algorithm  is built upon a multi-stage architecture that interleaves convolutions and max-pooling at three different scales among the feasible warpings between descriptors. The set of \textit{feasible warpings} $W$ is defined recursively  so that finding the optimal warping $w^*$ can be done efficiently by a dynamic programming strategy.
Fig. \ref{tracklet} illustrates an example of a generated tracklet based on deep matching for one of the seeds in the sequence shown in Fig.\ref{seed}. The seed is depicted by red bounding boxes, green bounding boxes correspond to the samples with highest deep matching score to the seed in every frame. As can be seen, the tracklet corresponds to the same person who generated the seed.


\subsection{Grouping tracklets into Bag-of-tracklets}
\label{grouping}
We assume that tracklets generated by seeds belong to the same person in the sequence, and are very likely to be similar to each other;  we aim to  group them into a set of eBoTs, where there is no intersection between eBoTs by definition.
Let us consider an eBoT, say $\mathbb{T}$, as a set containing a  tracklet, $\mathbb{T}=\{T^i\}$, where $T^i$  does not belong to any other eBoT. Also, let us consider another tracklet $T^j$ that has not been assigned to any eBoT yet.
Let $t_k^{i}$ and $t_k^{j}$ be the bounding boxes, where the person is detected (by the face detector or by the tracker) at frame $k$ for tracklets $T^i$ and $T^j$, respectively.

We define the similarity between two tracklets $T^i$ and $T^j$ as the average of the area of the intersection between  $t_k^{i}$ and $t_k^{j}$ divided by the area of their union:

$$\mathcal{S}(T^j,T^i)= \frac{1}{|T^i|}\sum_{k=1}^{|T^i|}\frac{|t_k^{j}\bigcap t_{k}^{i}|}{|t_{k}^{j}\bigcup t_{k}^{i}|}.$$

Given a tracklet $T^j$, it will be added to the eBoT $\mathbb{T}$, if the similarity between $T^j$ and all tracklets in $\mathbb{T}$ is high enough. In this work, we experimentally found that the threshold 0.2 to include  a tracklet in an eBoT provided the optimal results.
Before adding tracklets to an eBoT, we sort them based on their similarity to the first tracklet in the eBoT. Since the next tracklets need to be compared to the existing tracklets in an eBoT, sorting tracklets prior to other computations, helps avoiding aggregation of biased tracklets in the eBoT.

The similarity of tracklet, $T^j$ to the eBoT, $\mathbb{T}$ is defined as the average of the similarities to all its tracklets:
\begin{equation}
\mathcal{\widetilde{S}}(T^j,\mathbb{T})=\frac{1}{|\mathbb{T}|}\sum_{T^i\in \mathbb{T},T^i\neq T^j} \mathcal{S}(T^j,T^i)
\label{similarity_score}
\end{equation}
where $|\mathbb{T}|$ is the number of tracklets in the eBoT. After grouping by similarity, all tracklets in an eBoT are very likely to correspond  to the same  person.

However, not all tracklets in an eBoT are equally reliable. In addition, some eBoTs may correspond to seeds that are false positive detections. While the first issue is related to the prototype extraction and will be addressed in the next subsection, here we detail how to remove unreliable eBoTs that do not correspond to any person in the video.
To this end, we define the \emph{density} of an eBoT as  $d(\mathbb{T})=\frac{|\mathbb{T}|}{|T|}$, where $|\mathbb{T}|$  is the number of its tracklets and  $|T|$ is the length of the sequence.

\begin{figure}[H]
\begin{center}

{\includegraphics[width=11.8cm, height=10.7cm] {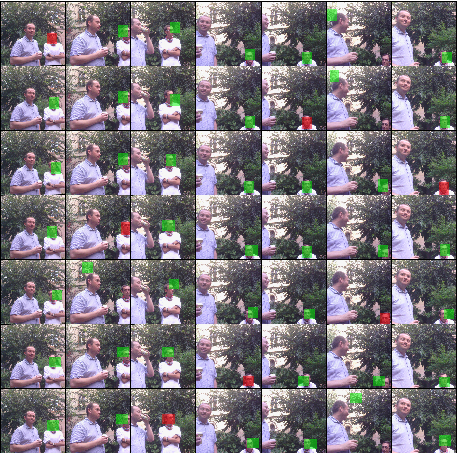}\label{ebot2}}
\caption{Example of a reliable eBoT -after excluding unreliable eBoT- extracted from the sequence in Fig. \ref{seed}. Each row shows a tracklet in the eBoT which totally consists of 7 tracklets. The red box in each row indicates the seed of that tracklet and green boxes to the samples with highest average deep matching score to their corresponding seed. As can be appreciated, all tracklets in the eBoT correspond to the same person.}
\label{ebot}
\end{center}
\end{figure}

Ideally, the density should be equal to 1 and we would have as much tracklets in the eBoT, as the number of frames the person persisted in the video. In practice, since the face detection algorithm as well as the matching algorithm may generate unreliable detections, the eBoT is looking for the consensus between the different tracklets to obtain the right tracking outcome. As expected, reliable eBoTs  show different behavior from unreliable ones, having the latter very low density. Based on this observation, we discard as unreliable all eBoTs having a density lower than a predefined threshold. In this work, we empirically found that a threshold of 0.2 gives good results. By excluding unreliable eBoTs, we obtain as number of eBoTs as the number of persons in that sequence (see Fig. \ref{ebot}).

\subsection{Prototype extraction}

A prototype extracted from an eBoT $\mathbb{T}$ should represent all tracklets in the eBoT. Thus, it should localize the face of a person in every frame. Since the detection of the target in a given frame of the sequence varies, depending on the seed that generated the tracklet, we choose as the prototype frame the one whose bounding box has the biggest intersection with the rest of the tracklets in that frame, namely:
$$
\hat{T}=\{\hat{t_b},\ldots,\hat{t_k},\ldots,\hat{t_e}\}, \mbox{ so that } 
 \hat{t_k}=\argmax_{ i=1,\ldots |\mathbb{T}|} \sum_{ j=1,...,|\mathbb{T}|, j \neq i} t_k^i \bigcap t_k^j,
$$
where $|\mathbb{T}|$ is the number of tracklets in the eBoT, $(t_k^i,t_k^j)_{i \neq j}$ are the bounding boxes of detected faces in the k-th frame of tracklets $T^i$ and $T^j$ from the eBoT $\mathbb{T}$, respectively.

\begin{figure}[h]
\begin{center}
\subfloat[]
{\includegraphics[width=11.8 cm, height=1.6 cm] {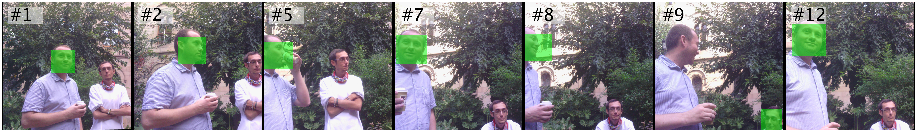}\label{subfig:correct}}

\subfloat[]
{\includegraphics[width=11.8 cm, height=1.6 cm] {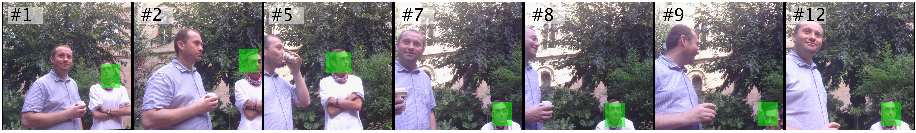}\label{subfig:correct}}
\caption{Two Prototypes extracted for the two persons in the sequence.}\label{prototype}
\end{center}
\end{figure}

Fig. \ref{prototype} shows two prototypes, each of them extracted from separate eBoT where only one of them is shown Fig. \ref{ebot}. Note how the prototype correctly tracks the person although the face detector misses the person in several frames. Missed detections can be seen in Fig. \ref{seed}. 

\subsection{Occlusion treatment}

Beside optimizing the localization of the target, a good prototype should also indicate the presence of occlusions or unreliable detections.
In order to increase the accuracy of the method, we  detect in the final prototype those frames, where the target is fully or partially occluded or there is an unreliable detection. To this goal, we define a function  $\Lambda( t_s^i,t^i_{k})$, that associates to each bounding box, $t^i_{k}$ of a tracklet $T^i$  the value of the deep matching score to its seed  $t^i_s$. We define a \textit{frame  confidence} as the average of the normalized deep matching scores of its bounding boxes of all the tracklets of the same eBoT, in that frame, that is:

\begin{equation}
\mathcal{C}_k=\frac{1}{|\mathbb{T}|} \sum_{i=1}^{|\mathbb{T}|} \Lambda (t_s^i,t_k^i),
\label{confidence1}
\end{equation}
In equation\ref{confidence1}, $\mathcal{C}_k$ is the frame confidence, $|\mathbb{T}|$ is the number of tracklets in the eBoT, $t_s^i$ is the seed of the i-th tracklet of the eBoT and $t_k^i$ is the bounding box of frame $k$ of the i-th eBoT tracklet. The deep matching scores between bounding boxes in the eBoT have been normalized between zero and one.

When there is a severe or partial occlusion of the face, or the target is missing, the confidence of the eBoT on that frame $\mathcal{C}_k$ experiences a drop. This phenomenon can be observed in Fig. \ref{prototypeconf}, where, due to partial occlusion of faces in frames 5 and 6 in Fig. \ref{prototypeconf} (a) and frames 6 in Fig. \ref{prototypeconf} (b), the confidence value in these frames has a minimum and lies under the pre-defined threshold for occlusion estimation. In all the cases of occlusions that are shown in Fig. \ref{prototypeconf} (a) and (b), the face of the person is only partially occluded. This fact shows the robustness of the method in estimating large changes in face appearance.

\begin{figure}[h]
\begin{center}
\includegraphics[width = \textwidth] {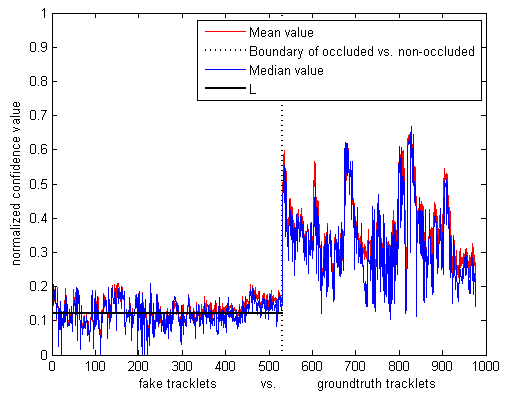}
\caption{Normalized confidence value for fake tracklets generated from an occluded target (left) and for groundtruth tracklets (right). The threshold $L$ we use to estimate occlusions is depicted in black.}\label{Threshold}
\end{center}
\end{figure}

\begin{figure}[H]
\begin{center}
\subfloat[]
{\includegraphics[width=12.5cm, height=5cm] {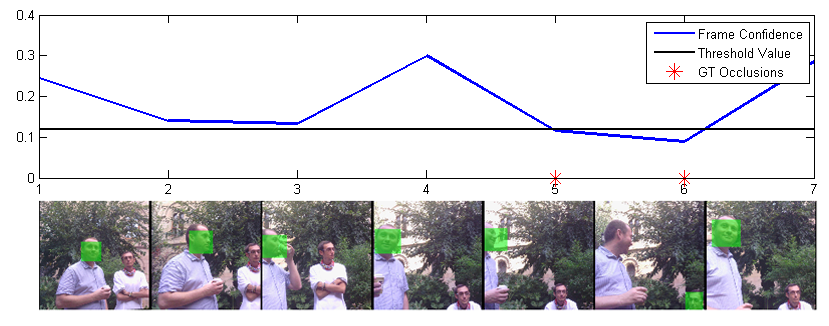}\label{subfig:correct}}

\subfloat[]
{\includegraphics[width=12.4cm, height=5cm] {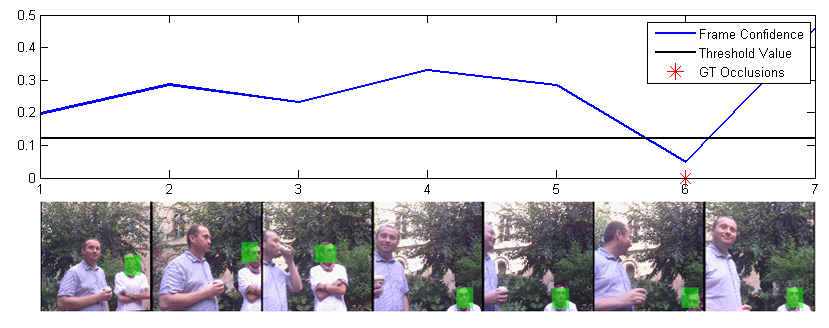}\label{subfig:correct}}
\caption{Frame confidence of two prototypes shown in Fig.\ref{prototype}, as defined in equation (\ref{confidence1}). The occurrence of occlusion for every person in the sequence in the groundtruth is shown by red stars in the plot. The black line corresponds to $L$, the threshold determined to estimate occlusions. As can be seen, the occurrence of the face occlusion indicated in the groundtruth, highly coincides with the calculated confidence drop of the face in that frame.}\label{prototypeconf}
\end{center}
\end{figure}

The value of the threshold for estimating occlusions, say $L$, is calculated over a subset of 15 sequences that constitute the training dataset. Fig. \ref{Threshold} shows the normalized confidence value calculated using equation \ref{confidence1}, for frames where the target is occluded (left) and for frames where the target is  not occluded (right). For non-occluded frames we used the groundtruth tracklet to compute the confidence values, whereas for occluded frames we generate a fake-tracklet by randomly defining a bounding box where there is not a face. As a tracklet is generated for each seed, in Fig. \ref{Threshold} we plot on the left the median value and the mean value of deep matching score over all the generated fake-tracklets and on the right the median value and the mean value of deep matching score over all the groundtruth tracklets over a sequence. The threshold $L$ (black line), emerges from the median of all the median confidence values over occluded frames.
We obtained this value as $L = 0.12$.


After estimating occlusions, we refine the frame confidence presented in equation \ref{confidence1}, considering it zero for occluded frames, that is:

\begin{equation}
\mathcal{C}_k= \left\lbrace 
\begin{array}{rl}
\frac{1}{|\mathbb{T}|} \sum_{i=1}^{|\mathbb{T}|} 
 \Lambda (t_s^i,t_k^i), & \mbox{ if } \frac{1}{|\mathbb{T}|}   \sum_{i=1}^{|\mathbb{T}|} \Lambda (t_s^i,t_k^i)\geq L
\\
0, & \mbox{ otherwise}
\end{array}
\right.
\label{confidence2}
\end{equation}

\subsection{Confidence of prototypes}

A prototype  can be very useful as a basis for  applications, such as finding type of a social interaction and social roles. Thus, confidence estimation of an extracted prototype is a valuable task. We define the \textit{prototype confidence} as the mean confidence over all its frames that do not undergo occlusion weighted by a term that penalizes occlusions, that is:

\begin{equation}
\begin{aligned}
\mathcal{C}(\hat{T}) = \frac{1}{|\hat{T}|}\sum_{k=1,\ldots, |\hat{T}|} \mathcal{C}(\hat{t_k}) \times max((1+\beta \log((|\hat{T}|-z)/|\hat{T}|)),0)
\end{aligned}
\label{prototype confidence}
\end{equation}
where  $|\hat{T}|$ is the length of the prototype, $z$ is the number of frames, where the face is occluded or missing, and $\beta$ is a control parameter that depends on the performance of the detector (we found that $\beta = 1$ gives reasonable results). Note that, in absence of occlusion, the confidence from equation (\ref{confidence2}) and equation (\ref{prototype confidence}) are the same.

Equation \ref{prototype confidence} is inspired from the definition of tracklet confidence given by Bae and Yoon in \textit{Multi-Object Tracking based on Tracklet Confidence} \cite{ILDA}. The first term is related to the coherence in appearance of the target along the tracklet: a more coherent appearance in a tracklet increases the confidence of the tracklet. The second term is related to the continuity of the tracklet: it decreases for occluded tracklets. Therefore, the final prototype should have a larger confidence than all the tracklets in an eBoT. After estimating occlusions for the prototypes, we associate a confidence value to each tracklet of the eBoT by using equation (\ref{prototype confidence}), and verify that the confidence of the prototype is higher than the highest tracklet confidence in the eBoT.  After evaluation, the average confidence value of all prototypes in our test set has a value of 0.54, which is higher than the average of the confidence value of all the tracklets in all eBoTs, being 0.32.


\section{Experiments and discussion}
\label{Results}

\subsection{Dataset}

Currently, there is no dataset for person tracking with groundtruth information in egocentric photo-streams. Hence, to measure the performance of the proposed model, we created a dataset acquired by the Narrative Clip camera\footnote{http://getnarrative.com/}. We manually annotated the sequences that contain trackable people and localized the position of their faces. The dataset has been acquired by five users of different ages. Each user wore the camera for a number of non-consecutive days over an 80 days period, collecting $\sim$20.000 images. Our dataset contains a total number of 108 different trackable persons along 80 sequences of average length of 25 frames\footnote{The dataset and the code will be made public domain, with the publication of the article.}. Table \ref{DataTable} provides further details of the proposed dataset.

\begin{table}[h]
\small
\caption{Detailed breakdown of our dataset made of  $\sim$20.000 images captured by  5 users}\label{table1}
\begin{center}
\begin{tabular}{| c | c | c | c | c | c | c |}
\hline
\textbf{User} & \textbf{Days} & \textbf{Total} & \textbf{Total frames} & \textbf{Total frames} & \textbf{Average daily} \\
\textbf{} & \textbf{} & \textbf{frames} & \textbf{with person(s)} & \textbf{with occlusion} & \textbf{duration}\\
\hline
1 & 30 & 6478 & 680 & 53 & 8h\\
2 & 5 & 1228 & 125 & 17 & 8h\\
3 & 10 & 3428 & 220 & 27 & 8h\\
4 & 28 & 6894 & 850 & 96 & 8h\\
5 & 7 & 2178 & 425 & 22 & 6h\\
\hline
\end{tabular}
\end{center}
\label{DataTable}
\end{table}


\subsection{Experimental setup}

After partitioning a photo-stream captured by the Narrative Clip into segments, a face detector is applied to exclude non-trackable segments and generate possible seeds for trackable segments, called sequences. Then, a tracklet is generated for each seed in a sequence. Finally, the tracklets are grouped into eBoTs and a final prototype with estimated occlusion is extracted from each reliable eBoT. These prototypes constitute the final output of our method. In the next section, quantitative and qualitative comparison between our approach and other tracking approaches is provided.

We measured the performance of our method by using the CLEAR MOT \cite{metrics}  on the resulting prototypes (with and without occlusion  estimation). Additionally, we compared its performance with other six state of the art methods.  CLEAR MOT consists of multiple metrics as follows. The Multiple Object Tracking Precision (MOTP) evaluates the intersection area over the union area of the bounding boxes:
$$
MOTP = \frac{1}{|M_s|}\sum_{k\in  M_s}\frac{|t_k \bigcap gt_k|}{|t_k \bigcup gt_k|},
$$
\noindent where $M_s$ is the set of frames in a sequence in which the tracked bounding box $t_k$ intersects  the groundtruth bounding box $gt_k$, and $|M_s|$ is the cardinality of $M_s$. MOTP quantifies the accuracy of the tracker by estimating the precise location of the object, regardless of its ability in keeping consistent trajectories.

On the other side, the Multiple Object Tracking Accuracy (MOTA)  estimates the accuracy of the results by penalizing False Negatives (FN), False Positives (FP) and IDentity Switching (IDS), namely:
$$
MOTA = 1 - \frac{\sum_{k=1}^l (FN_k+FP_k+IDS_k)}{\sum_{k=1}^l GT_k},
$$
where $k$ refers to the frame number, $l$ is the length of the sequence, and $GT_k$ states for the number of faces in the ground-truth to be tracked at frame $k$. 
$FN_k$ and  $FP_k$ donate the number of false negatives and false positives in a frame $k$, respectively. $IDS_k$ is equal to 1 when the detection does not overlap with its corresponding groundtruth face target, but with another face. 

Both metrics intuitively express the overall strength of each tracker and are suitable for general performance evaluations. Furthermore, the qualitative comparative results are also shown over four different sequences in the next section.

\subsection{Discussion}


\textbf{Quantitative evaluation:}  
To the best of our knowledge, the only work which is exclusively introduced for person tracking in egocentric photo-streams is  BoT \cite{BOT}. Most of the available tracking techniques are not directly applicable to egocentric photo-streams, since they follow assumptions such as temporal consistency between frames or smooth variation in target and background appearance, that do not hold for egocentric photo-streams. Still, we compared our approach to six different state of the art algorithms that 
are applicable to egocentric photo-streams, since they do not rely on motion information nor background modeling. The selected trackers are designed for tracking one object at time, but in our dataset more than one person appears in the sequence. Thus, we applied the trackers separately for each person to adapt them to our scenario. In this case, the tracking problem reduces to one object tracking and therefore for evaluation measurements we do not consider the IDS metric for these methods as proposed by Smeulders et al. in \cite{survey}.
In Table \ref{table2}, we show the percentage of MOTP, MOTA, FP, FN and IDS on the results of AMT \cite{NNLFR}, BoT \cite{BOT}, CT \cite{CT}, LOT \cite{LOT}, L1O \cite{L1O}, and SPT \cite{SPT}. We also show how the estimation of occlusions improves the performance of the proposed method in most of the metrics. 

\begin{table}[h]
\caption{Performance comparison}\label{table2}
\resizebox{\textwidth}{!}{%
\begin{tabular}{l*{6}{c}r}
\hline
\hline

Methods              & MOTP$\uparrow$ & MOTA$\uparrow$ & FP$\downarrow$ & FN$\downarrow$ & IDS$\downarrow$ \\
\hline
AMT (Abrupt Motion Tracking)  & 60.99\% & 59.65\% & 16.70\% & 23.65\% & - \\
BoT (Bag of Tracklets) & 48.39\% & 43.44\% & 22.9\% & 20.17\% & 14.30\% \\
CT (Compressive Tracking)  & 35.05\% & 15.32\% & 33.07\% & 51.61\% & - \\
LOT (Locally Orderless Tracking) & 42.27\% & 15.57\% & 33.12\% & 51.13\% & - \\
L1O (L1 Tracker with Occlusion Detection ) & 37.25\% & 25.87\% & 31.81\% & 42.32\% & - \\
SPT (SuperPixel Tracking) & 40.75\% & 39.31\% & 23.56\% & 37.13\% & - \\
\hline

eBoT (prototype, occlusions not excluded) & 68.32\% & 72.08\% & 15.19\% & 10.60\% &  2.13\% \\
eBoT (prototype, occlusions excluded) & 70.27\% & 80.23\% & 5.12\% & 12.51\% &  2.13\% \\
\hline
\hline
\label{ResultTable}
\end{tabular}}
\end{table}

As can be observed, the difference among CT, LOT, L1O, and SPT in terms of precision (MOTP) is small, where CT has the smallest value. This can happen, since this tracker does not change the scale of the bounding box, while other methods have a relatively good mechanism of scale adaptation. BoT and AMT have higher precision than other methods, being AMT that outperforms BoT. This can be justified in the way that in AMT the true object is introduced for the tracker in the initial frame of the sequence, whereas BoT is fully automatic.

In terms of accuracy (MOTA), CT and LOT performs much the same as each other. This might be a consequence of the fact that regular appearance model updates for both  trackers, thus they fail when they encounter a large variation between frames. However, L1O and SPT perform slightly better, since they are able to estimate occlusions, leading  to lower amount of FPs. SPT and LOT use superpixels representation, which is more suited for bigger objects. Thus, they perform better, when the face is closer to the camera and looks bigger. On the other hand, AMT is designed for tracking on low frame rate videos and performs quite good on our dataset, being able to outperform BoT. However, it can easily miscalculate the position of the target, when there are more than one face in the frame. The miscalculation may happen due to use of a color-based likelihood model that can easily get misled by finding a region with similar colors to the target.

As one can see in the lower part of the Table \ref{table2}, the method proposed in this paper performs much better than the state of the art. The seventh and the eight lines in the Table \ref{table2} show evaluation metrics obtained before and after estimating occlusions.
The estimation of occlusions  allows to reduce FP, while slightly increases the FN rate due to wrongly eliminating some true detections in the final prototypes. The proposed method for prototype extraction allows to drastically reduce FP, FN and IDS, since it optimizes the localization of the detection.

From this evaluation, we can state that the proposed system can robustly track multiple person's face under challenging conditions. Moreover, this improvement is achieved without relying on any strong assumptions and without the need of a cumbersome training stage.

\begin{figure}[h]
\begin{center}
\subfloat[]
{\includegraphics[width=11.8cm, height=1.6cm] {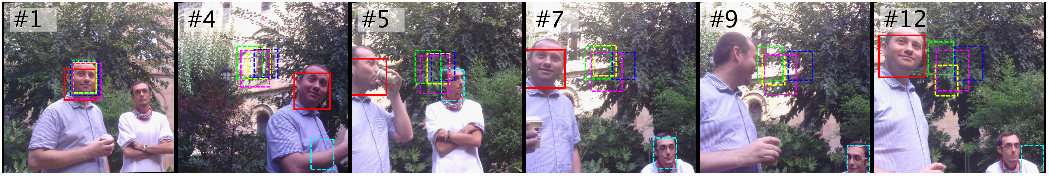}\label{subfig:correct}}

\subfloat[]
{\includegraphics[width=11.8cm, height=1.6cm] {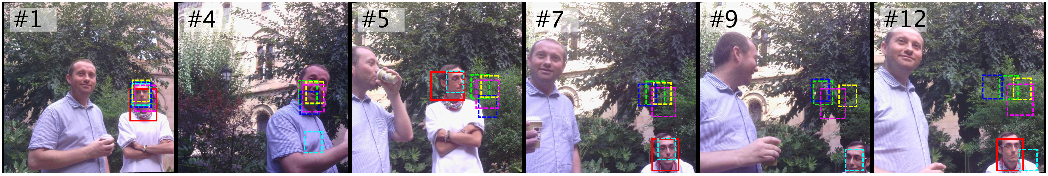}\label{subfig:correct}}
\caption{Results of applying different methods on an egocentric photo-stream. Different bounding boxes show the tracking results of the \textcolor{green}{CT}, \textcolor{Blue}{LOT}, \textcolor{Cyan}{AMT}, \textcolor{yellow}{SPT}, \textcolor{magenta}{L1O} and \textcolor{red}{our} proposed approach. Occlusions can be observed in frame \#9 (a) and frames \#4 and \#9 (b).}
\label{exp63}
\end{center}
\end{figure}

\begin{figure}[H]
\begin{center}
\subfloat[]
{\includegraphics[width=11.8cm, height=1.6cm] {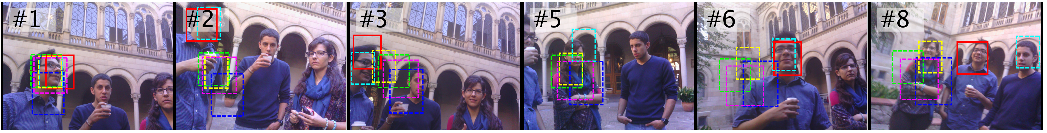}\label{}}

\subfloat[]
{\includegraphics[width=11.8cm, height=1.6cm] {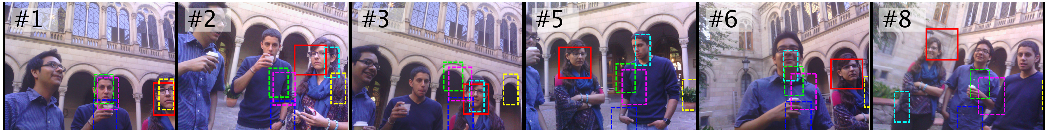}\label{}}

\subfloat[]
{\includegraphics[width=11.8cm, height=1.6cm] {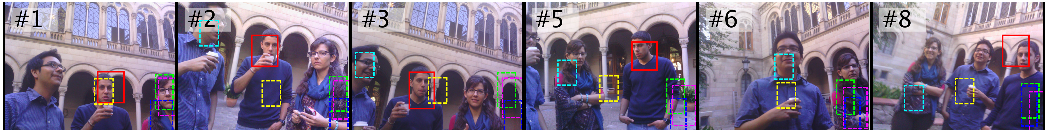}\label{}}
\caption{Results of applying different methods on an egocentric photo-stream. Different bounding boxes show the tracking results of the \textcolor{green}{CT}, \textcolor{Blue}{LOT}, \textcolor{Cyan}{AMT}, \textcolor{yellow}{SPT}, \textcolor{magenta}{L1O} and \textcolor{red}{our} proposed approach. Occlusions can be observed in frame \#5 (a) and frame \#6 (c).}
\label{exp64}
\end{center}
\end{figure}

\begin{figure}[h]
\begin{center}
\subfloat[]
{\includegraphics[width=11.8cm, height=1.6cm] {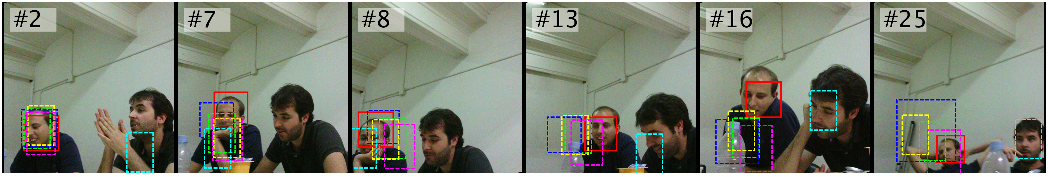}\label{}}

\subfloat[]
{\includegraphics[width=11.8cm, height=1.6cm] {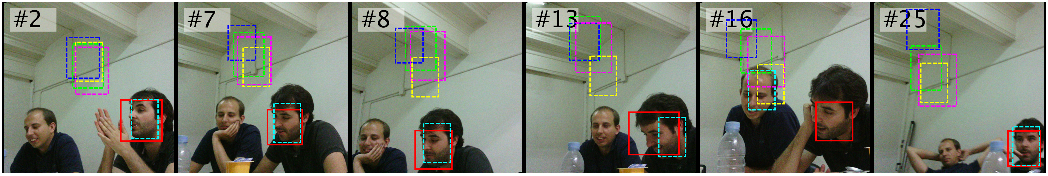}\label{}}
\caption{Results of applying different methods on an egocentric photo-stream. Different bounding boxes show the tracking results of the \textcolor{green}{CT}, \textcolor{Blue}{LOT}, \textcolor{Cyan}{AMT}, \textcolor{yellow}{SPT}, \textcolor{magenta}{L1O} and \textcolor{red}{our} proposed approach.}\label{exp49}
\end{center}
\end{figure}

\begin{figure}[h]
\begin{center}
\subfloat[]
{\includegraphics[width=11.8cm, height=1.6cm] {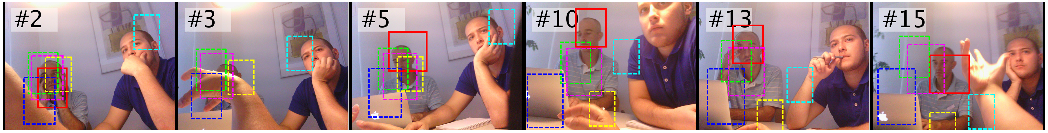}\label{}}

\subfloat[]
{\includegraphics[width=11.8cm, height=1.6cm] {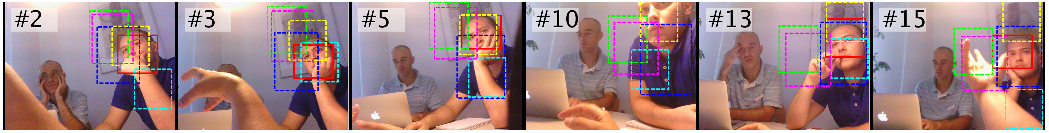}\label{}}
\caption{Results of applying different methods on an egocentric photo-stream. Different bounding boxes show the tracking results of the \textcolor{green}{CT}, \textcolor{Blue}{LOT}, \textcolor{Cyan}{AMT}, \textcolor{yellow}{SPT}, \textcolor{magenta}{L1O} and \textcolor{red}{our} proposed approach. Occlusions can be observed in frame \#3 (a) and frame \#10 (b).}\label{exp26}
\end{center}
\end{figure}

\textbf{Qualitative evaluation:}
The tracking results of the proposed approach together with the results of previously introduced trackers is shown over four different sequences in Fig. \ref{exp63}, Fig. \ref{exp49}, Fig. \ref{exp26}, and Fig. \ref{exp64}. Every sequence contains multiple persons and tracking result of each tracker is shown by a specific color in every frame of the sequences. The result of the proposed approach is shown by a red bounding box around the face of the person. In the frame, where our method detects an occlusion, no bounding box is shown. For the sake of visualization, if a sequence contains more than one person, the tracking result for each person is shown in a separate line. 
Fig. \ref{exp63} shows the final prototypes with estimated occlusions of the prototypes shown in Fig. \ref{prototype}. Fig. \ref{exp49} and Fig. \ref{exp26} show the result for a sequence of two different persons and Fig. \ref{exp64} shows them for a sequence of three different persons. 

Among the state of the art methods, AMT has the best performance on our dataset, because it was designed to cope with abrupt motion changes. However, it can easily produce FPs in presence of multiple persons for not being a multi-tracking method. As can be observed, CT, LOT, L1O, and SPT are disable to find the target, when its location varies largely. In addition, a common drawback among the AMT, BoT, CT, and LOT is that they are unable to localize target occlusion.
As expected, it can be seen that the tracking results of the proposed approach highly match the person face. However, the method assigns a wrong region to the track, when a person face is occluded, causing the occurrence of FPs or IDS. Still, our method is able to precisely estimate occlusions or wrongly assigned detection.

From our experiments, we could observe  that the proposed method works better, when the people are closer to the camera. As the distance of the people from the camera increases, the resolution of the image on their face region decays. That phenomenon leads to generation of less seeds by the face detector and to unreliable matches by the deep matching approach. The illumination condition is another important factor as well. eBoTs is quite robust to illumination changes, although it performs better, when the images are not too dark.

\subsection{Complexity analysis}
Regarding the complexity of our algorithm, one can easily see that the most expensive part is the construction of the tracklets, where the deep matching is applied with a sliding window procedure to all windows having a similar color to the seed in the HSV color space. The most expensive part of the deep matching algorithm lies in the computation of the first level convolutions. However, the computational burden would be mitigated by using a GPU or a faster matching algorithm \cite{fastmatch}, that achieves similar performances. Finding the optimal matching score among all feasible non-rigid warpings  for all square patches at different scales, from the first image at all locations in the second image can be done with complexity $O(PP')$, where $P$ and $P'$ are the number of pixels of both images. Usually, the size of the seed image is between 5000 and 6000 pixels and the number of samples to be considered is about 2000.
On a CPU Intel i5 - 2.53 GHz, with operating system Windows 7 - 64 bit, 4G of RAM, it takes in average about 1 minutes per each pair of images to find the similar candidate to the seed. It is easy to see that the complexity of the rest of algorithms to construct the eBoT and extract the prototype is $O(M*N^2)$, where $M$ is the number of faces appearing in the sequence and $N$ is the length of the sequence, taking less than a minute in the aforementioned computer. 


\section{Conclusions}
\label{sec:conclusions}
In this work, we proposed a novel method to track multiple-faces in low temporal resolution sequences acquired by wearable cameras, that is of high interest to analyze social events and social interactions in egocentric vision. Relying on the {\textcolor{red} extended bag-of-tracklets approach 
for tracking a person increases the robustness and efficiency of our method. To deal with various types of object-induced or camera-induced image deformations, tracklets are computed by using the average deep-matching score between the seed and each sample in different frames. Moreover, in order to extract the final prototype, eBoT introduces a useful measure of confidence to estimate and discard occlusions and missed detections.

A quantitative comparison of in a dataset of 20.000 images between our model and other six state of the art methods showed its advantage  under drastic changes of poses, scales and object appearances.
Future work will be devoted to quantify the kind of interaction with the camera wearer as well as to detect and classify social events. Human memories are influenced by emotions and strong emotional impact of social interaction is well acknowledged. Thus, a direct application will be to use the extracted prototypes for cognitive training of patients with mild cognitive impairment.


\section*{Acknowledgments}
This work was partially founded by projects TIN2012-38187-C03-01 and SGR 1219. The first author is supported by an \textit{APIF} grant from the University of Barcelona.
The second author is supported by a \textit{Beatriu de Pin\`os} grant (Marie-Curie COFUND action). The third author is partly supported by an \textit{ICREA Academia} grant. 


\bibliographystyle{elsarticle-num}

\bibliography{main}

\end{document}